%% file: 360feedback.tex
\pdfoutput=1
\documentclass[11pt]{article}
\usepackage{booktabs}
\usepackage{graphicx} 
\usepackage{algpseudocode}
\usepackage[final]{acl}
\usepackage{multirow}
\usepackage[inline]{enumitem}
\usepackage{enumerate}  
\usepackage{amsfonts}
\usepackage{boldline}
\usepackage{bm}
\usepackage{latexsym}
\usepackage{xspace}
\usepackage{amsmath}
\usepackage{pifont}
\usepackage{amssymb}
\usepackage{newfloat}
\usepackage{listings}
\usepackage{times}
\usepackage{latexsym}
\usepackage[T1]{fontenc}    
\usepackage[utf8]{inputenc}
\usepackage[dvipsnames]{colortbl}
\usepackage[T1]{fontenc}
\usepackage{babel} 
\usepackage[toc,page]{appendix}
\usepackage{makecell}
\usepackage{soul}
\usepackage{verbatim}
\usepackage{amssymb}
\usepackage{tcolorbox}

\usepackage{microtype}

\usepackage{inconsolata}
\definecolor{lemon}{HTML}{FDFFCC}

\newcommand{\instrctionsize}{\footnotesize}
\newcommand{\model}{$360^\circ$REA\xspace}
\newcommand{\fullmodel}{\textbf{R}eusable \textbf{E}xperience \textbf{A}ccumulation with $\mathbf{360^\circ}$ Assessment\xspace}
\newcommand{\revm}{$360^\circ$ performance assessment\xspace}
\newcommand{\expm}{dual-level experience pool\xspace}

\definecolor{Gainsboro}{rgb}{0.86, 0.86, 0.86}

\definecolor{Gray}{gray}{0.95}
\definecolor{LightCyan}{rgb}{0.88,1,1}

\newcommand{\paratitle}[1]{\noindent\textbf{#1}}

\newcommand{\aka}{\emph{a.k.a.,}\xspace}
\newcommand{\eg}{\emph{e.g.,}\xspace}

\newcommand{\ignore}[1]{}

\title{\model: Towards A Reusable Experience Accumulation with $360^\circ$ Assessment for Multi-Agent System}

\author{
Shen Gao$^{1}$\footnotemark[2], Hao Li$^{2}$\footnotemark[2], Chengrui Huang$^{1}$,\\
{\bf Quan Tu$^{3}$, Zhiliang Tian$^{4}$\textsuperscript{*}, Minlie Huang$^{5}$, Shuo Shang$^{1}$} \\
$^{1}$ University of Electronic Science and Technology of China,\\
$^{2}$ Shandong University,
$^{3}$ Renmin University of China,\\
$^{4}$ National University of Defense Technology,
$^{5}$ Tsinghua University \\ 
\texttt{shengao@pku.edu.cn},
\texttt{lihaohan@mail.sdu.edu.cn},\\
\texttt{\{yunrongyuxi, jedi.shang\}@gmail.com},
\texttt{quantu@ruc.edu.cn},\\
\texttt{tianzhiliang@nudt.edu.cn},
\texttt{aihuang@tsinghua.edu.cn}
}

\begin{document}
\maketitle
\footnotetext[2]{These authors contributed equally to this work.}
\renewcommand{\thefootnote}{\arabic{footnote}}
\begin{abstract}
Large language model agents have demonstrated remarkable advancements across various complex tasks.
Recent works focus on optimizing the agent team or employing self-reflection to iteratively solve complex tasks.
Since these agents are all based on the same LLM, only conducting self-evaluation or removing underperforming agents does not substantively enhance the capability of the agents.
We argue that a comprehensive evaluation and accumulating experience from evaluation feedback is an effective approach to improving system performance.
In this paper, we propose \fullmodel (\model), a hierarchical multi-agent framework inspired by corporate organizational practices. 
The framework employs a novel \revm method for multi-perspective performance evaluation with fine-grained assessment.
To enhance the capability of agents in addressing complex tasks, we introduce \expm for agents to accumulate experience through fine-grained assessment. 
Extensive experiments on complex task datasets demonstrate the effectiveness of \model\footnote{Code is available at \url{https://github.com/LLLQAQFFF/360REA.git}}. 
\end{abstract}

\input{sections/01-introduction}
\input{sections/02-related-work}
\input{sections/04-methodology}
\input{sections/05-experiment}

\input{sections/06-result-and-analysis}
\input{sections/08-conclusion}

\bibliography{custom}
\clearpage
\appendix
\input{sections/09-appendix}

\end{document}

%% file: sections/01-introduction.tex
\section{Introduction}~\label{sec:intro}


A variety of tasks, spanning from reasoning~\cite{bestaGraphThoughtsSolving2023,wangSelfConsistencyImprovesChain2023,weiChainofThoughtPromptingElicits2023,yaoREACSYNERGIZINGREASONING2023,yaoTreeThoughtsDeliberate2023,zhouLeasttoMostPromptingEnables2023} to code generation~\cite{anishkaCanChatGPTPlay2023,chenCodeTCodeGeneration2022,chenEvaluatingLargeLanguage2021,dongSelfcollaborationCodeGeneration2023a,hongMetaGPTMetaProgramming2023,wuAutoGenEnablingNextGen2023}, as well as video gaming~\cite{wangVoyagerOpenEndedEmbodied2023,zhuGhostMinecraftGenerally2023} and autopilot~\cite{dongSelfcollaborationCodeGeneration2023a,lykourentzouSelfOrganizingTeamsOnline2021,nascimentoGPTintheLoopAdaptiveDecisionMaking2023}, have seen promising advancements through the utilization of Large Language Model (LLM) agents. 
Recent works have explored the effectiveness of combining multiple LLM agents to collaborate on complex tasks~\cite{chenAgentVerseFacilitatingMultiAgent2023,liuBOLAABenchmarkingOrchestrating2023,packerMemGPTLLMsOperating2023,ren2024healthcare,DBLP:conf/coling/WangTSZPTH024}. 
These methods can generally be classified into two main categories. 
One category involves pre-defining the roles and specific tasks of each agent within a multi-agent system, such as software development~\cite{hongMetaGPTMetaProgramming2023,qianCommunicativeAgentsSoftware2023}. 
The other category of methods allows multiple agents to decide their tasks and roles autonomously, \eg travel planning~\cite{zhengAgentsMeetOKR2023}. 
In these methods, they usually focus on how to allocate tasks among agents~\cite{chengCOOPERCoordinatingSpecialized2023,wangUnleashingCognitiveSynergy2023} or how to score agents to subsequently remove those with low scores~\cite{liuDynamicLLMAgentNetwork2023}. 
While some methods~\cite{xiSelfPolishEnhanceReasoning2023,shinnReflexionLanguageAgents2023,wangSelfInstructAligningLanguage2023} aim to correct the result of the current task by employing self-reflection in the agent, these approaches are limited to improving current specific task performance. 
The capability of the agent to address similar tasks is not enhanced.

In a company, the ultimate goal of company management is not merely to score employee performance, to promote or fire them.
The best practice is to use the performance assessment to assist employees in purposefully reflecting on their work, thereby enhancing their capability to fulfill their roles better.
360-degree assessment~\cite{Bracken2016TheEA,alimo1998360,peiperl2001getting} is a comprehensive process that involves collecting assessment on an employee's behavior from various perspectives, including colleagues, supervisors, and sometimes external sources like customers. 
Business organizations are increasingly incorporating 360-degree assessment into performance evaluations and employment decisions, such as pay and promotions~\cite{feishu}.
Motivated by this, in multi-agent systems, the design principle should be to assist agents in accumulating experience based on their assessment results and enabling them to perform better in subsequent tasks. 
This principle emphasizes the importance of helping agents to improve rather than solely evaluating or removing underperforming agents from the system. 
Therefore, designing a mechanism for agent assessment and capabilities learning poses a challenge in the design of multi-agent systems.

In this paper, we propose \fullmodel (\model) which is a hierarchical multi-agent interaction framework.
The objective is to enhance performance in handling complex tasks by emulating organizational structures, employee performance assessments, and the accumulation of experience within a company. 
We employ a hierarchical structure for organizing agents, where the leader agent allocates tasks and roles to crew agents, and the crew agents collaborate to accomplish the given task. 
To evaluate the performance of each crew agent, instead of using only self-reflection, we introduce a novel \textbf{\revm} method that employs a multidimensional evaluation in crew-crew and leader-crew dimensions. 
It provides agents with more valuable assessment by evaluating from both the peer level and supervisory level.

As mentioned above, the accumulation of experience is essential to achieve better agent performance~\cite{madaanSelfRefineIterativeRefinement2023,miaoSelfCheckUsingLLMs2023,shinnReflexionLanguageAgents2023,wangSelfInstructAligningLanguage2023}. 
In this paper, we propose a \textbf{\expm} to help agents collect useful experiences when tackling complex tasks. 
Specifically, within the \textit{local} experience pool, agents utilize fine-grained insights from the \revm to reflect on their current outputs and form experiences related to the sub-tasks. 
To effectively address a complex task, optimizing individual sub-tasks separately is sub-optimal, and it is essential to a comprehensive understanding and analysis of the complex task.
To construct a high-level experience, leader agents use the evaluation of the final result to comprehensively summarize the task, forming a \textit{global} experience pool. 
Finally, the team accumulates sufficient experience in handling such tasks, leading to improved performance on similar tasks.
Extensive experiments conducted on two widely used complex task datasets demonstrate the effectiveness of our proposed \model.

\input{figures/comp}

\noindent Our contributions of this work are as follows:

\noindent $\bullet$ We propose \model, which is a multi-agent interaction framework inspired by contemporary corporate organizational governance methods.

\noindent $\bullet$ We propose \revm to evaluate the performance of agents from multiple dimensions.

\noindent $\bullet$ We introduce \expm to help agents collect useful experiences based on the evaluation result.

\noindent $\bullet$ Experimental results on benchmark datasets illustrate the advantage of \model on several complex tasks.

%% file: figures/comp.tex
\begin{table}[htbp]
\centering
\resizebox{1.0\columnwidth}{!}{
\begin{tabular}{lccccc}
\toprule
\textbf{Method}
& \begin{tabular}[c]{@{}c@{}} \textbf{Prompting} \\\end{tabular}
&  \begin{tabular}[c]{@{}c@{}} \textbf{SPP} \\  \end{tabular}   
& \begin{tabular}[c]{@{}c@{}} \textbf{DyLAN} \\\end{tabular}   
& \begin{tabular}[c]{@{}c@{}} \textbf{OKR} \\\end{tabular}
& \begin{tabular}[c]{@{}c@{}} \textbf{\model} \\\end{tabular}   
\\

\cmidrule(r){1-1}
\cmidrule(r){2-2}
\cmidrule(r){3-3}
\cmidrule(r){4-4}
\cmidrule(r){5-5}
\cmidrule(r){6-6}

Hierarchical Agents
& \textcolor{red}{\ding{56} }
& \textcolor{red}{\ding{56} }
& \textcolor{red}{\ding{56} }
& \ding{52}
& \ding{52}
\\

Dynamic Task Allocation
& \textcolor{red}{\ding{56} }
& \ding{52}
& \textcolor{red}{\ding{56} }
& \ding{52}
& \ding{52}
\\

Peer Assessment
& \textcolor{red}{\ding{56} }
& \textcolor{red}{\ding{56} }
& \ding{52}
& \textcolor{red}{\ding{56} }
& \ding{52}
\\


Self Assessment
& \textcolor{red}{\ding{56} }
& \textcolor{red}{\ding{56} }
& \ding{52}
& \ding{52}
& \ding{52}
\\

Supervisory Assessment
& \textcolor{red}{\ding{56} }
& \textcolor{red}{\ding{56} }
& \textcolor{red}{\ding{56} }
& \textcolor{red}{\ding{56} }
& \ding{52}
\\

Reusable Experience
& \textcolor{red}{\ding{56} }
& \textcolor{red}{\ding{56} }
& \textcolor{red}{\ding{56} }
& \textcolor{red}{\ding{56} }
& \ding{52}
\\

Agent Discussion
& \textcolor{red}{\ding{56} }
& \ding{52}
& \textcolor{red}{\ding{56} }
& \textcolor{red}{\ding{56} }
& \ding{52}
\\

\bottomrule
\end{tabular}
}
\caption{Comparison of existing multi-agent collaboration frameworks. }
\label{tab:comparison}
\end{table}

%% file: sections/02-related-work.tex
\section{Related Work}~\label{sec:related-work}

Large-scale language models (LLMs) such as ChatGPT have demonstrated outstanding performance across various natural language tasks~\cite{openaiGPT4TechnicalReport2023}. 
In recent years, several methods have emerged to enhance the ability of LLM to tackle complex tasks, such as Chain of Thought~\cite{weiChainofThoughtPromptingElicits2023}, Tree of Thoughts~\cite{yaoTreeThoughtsDeliberate2023}, and Graph of Thoughts~\cite{bestaGraphThoughtsSolving2023}.
These methods aim to improve reasoning by simulating human thinking processes within LLMs.
ReAct~\cite{yaoREACSYNERGIZINGREASONING2023} enables LLMs to generate interpretable action plans before taking actions, thus facilitating the completion of more complex tasks. 
Reflection~\cite{shinnReflexionLanguageAgents2023} enhances task performance by encouraging LLM to engage in self-reflection.
However, some complex real-world tasks typically require collaboration among multiple people to accomplish. 
Thus, relying solely on prompting methods often yields suboptimal performance~\cite{xiSelfPolishEnhanceReasoning2023,weiChainofThoughtPromptingElicits2023,yinExchangeofThoughtEnhancingLarge2023}. 

Recent work has begun exploring frameworks involving multiple LLM-based agents, which significantly outperform direct prompting methods in natural language understanding and reasoning tasks~\cite{chenAgentVerseFacilitatingMultiAgent2023,hongMetaGPTMetaProgramming2023,chenReConcileRoundTableConference2023}.
Some work involves conducting several rounds of debate among multiple LLM-based agents to enhance their authenticity and reasoning abilities~\cite{liangEncouragingDivergentThinking2023}. 
Other works propose dividing tasks into subtasks and employing multiple LLM-based agents to address complex tasks through a divide-and-conquer approach~\cite{chenAgentVerseFacilitatingMultiAgent2023,hongMetaGPTMetaProgramming2023,chenAutoAgentsFrameworkAutomatic}.
Furthermore, some works employ scoring the performance of multiple LLM-based agents and terminating those with poor performance to better accomplish tasks~\cite{liuDynamicLLMAgentNetwork2023}.

However, existing methods for evaluating agent performance often rely on the self-reflections of agents, neglecting comprehensive considerations within the entire multi-agent framework.
And solely evaluating and terminating low-scoring agents does not substantially enhance their task-solving capabilities. 

%% file: sections/04-methodology.tex
\section{\model Methodology}\label{sec:method}

\begin{figure*}[htbp]
 \centering
 \includegraphics[width=0.9\linewidth]{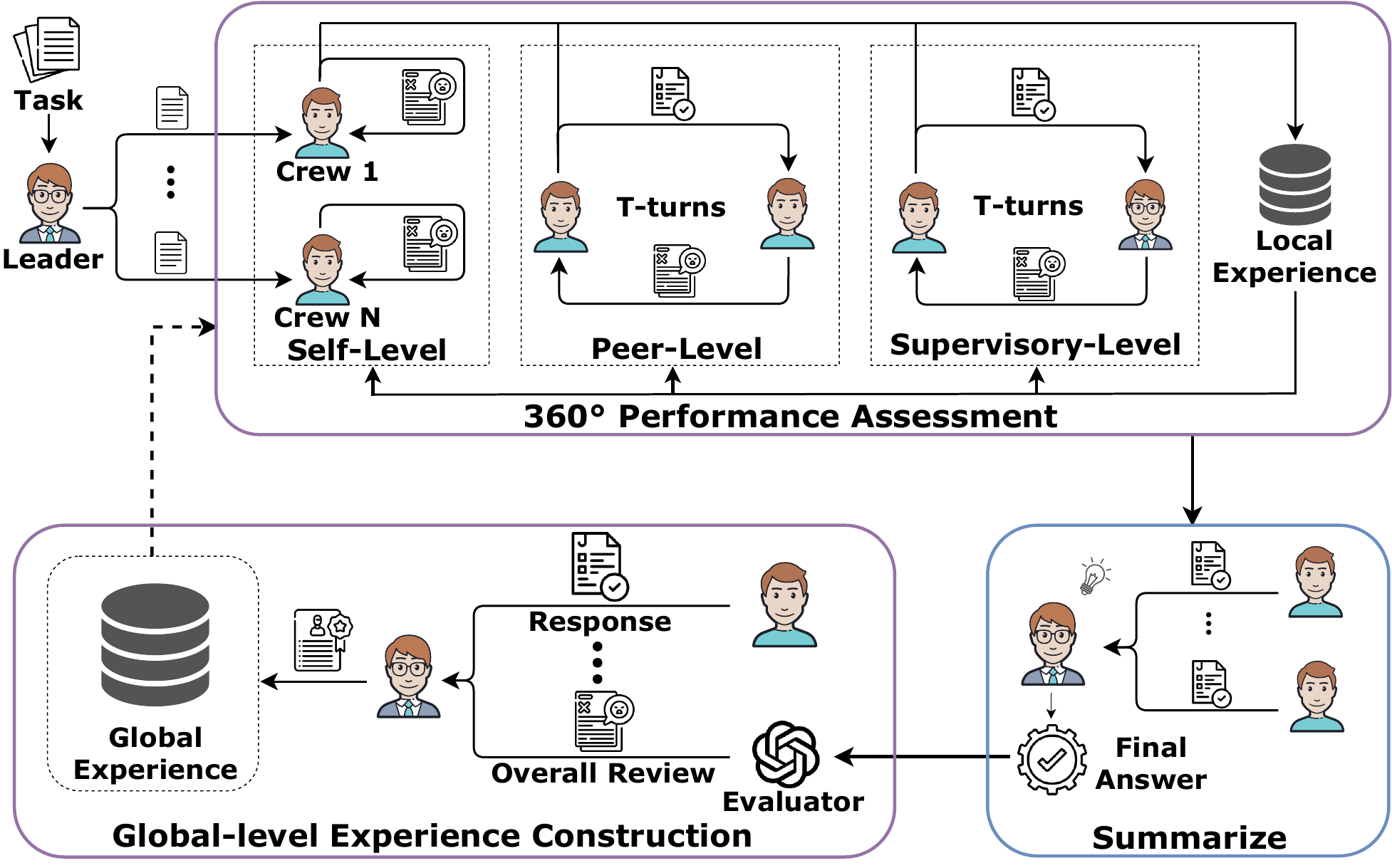}
 \caption{Our proposed \model. There are two main parts in our \model, including the \textbf{$360^\circ$ performance assessment} and \textbf{dual-level experience pool}. $360^\circ$ performance assessment can assist the agent in obtaining evaluations from multiple aspects. These evaluations will facilitate agents in refining their results and accumulating reusable experiences for accomplishing tasks better. Then, we store low-level specific and higher-level experiences in local and global experience pools separately.}
 \label{fig:model}
\end{figure*}

\subsection{Overview}\label{sec:overview}

In this section, we detail the \fullmodel (\model).
An overview of \model is shown in Figure~\ref{fig:model}.
We will detail the \model in three parts:
\begin{enumerate*}[label=(\roman*)]
\item We first illustrate the main workflow of the \model which uses the user query $x$ as input and generates the response $y$.
\item Next, we introduce the \revm which prompts the agents to give a review of each other and update their response according to others' reviews.
\item After finishing the response generation of each agent, each agent will summarize their experience and construct a dual-level pool that contains reusable task-solving experiences.
\end{enumerate*}

\subsection{Main Workflow of \model}

In this paper, we propose the \model which is a general multi-agent learning framework and can be applied to the hierarchical LLM-based multi-agent frameworks~\cite{zhengAgentsMeetOKR2023}.
In this LLM-based multi-agents framework, it consists of a leader agent $A^l$ and $N$ crew agents $\{A^c_1, A^c_2, \dots, A^c_N\}$ to solve the given task defined by the user input query $x$.
Given an input query $x$, the leader agent $A^l$ first generates the instruction of each crew agent, which first divides the task into several sub-tasks:
\begin{equation}
    \{I^c_1, I^c_2, \dots, I^c_N\} = A^l(x),
\end{equation}
where $I^c_i$ is the task instruction for the $i$-th crew agent.

To enhance the crew agent with task-solving capabilities, we propose \expm (introduced in \S~\ref{sec:exp-pool}), which contains to sub-experience pool: global experience $E_g$ and local experience $E_{l,i}$.
The experiences in \expm describe (1) the general skill of tackling the task and (2) the specific skill of the sub-task for agent $A^c_i$ respectively.
Then the crew agents generate their responses according to the instruction and useful context to solve the sub-task.
\begin{equation}
    H^t_i = A^c_i(I_i, E_g, E_{l,i}, R^t_{i}),\label{eq:crew-gen}
\end{equation}
where $H^t_i$ is the response of the $i$-th agents in $t$-th turn.
In this paper, we propose a multi-turn response generation paradigm for the crew agents.
After each turn of the agent generating the response $H^t_i$, we employ the \revm (introduced in \S~\ref{sec:360}) to evaluate the performance of agent $A^c_i$ from multiple perspectives and generate a review set $R^t_{i}$.
In Equation~\ref{eq:crew-gen}, we also incorporate the reviews from multiple sources $R^t_{i}$ when generating the response of the agent.
By this mechanism, agent $A^c_i$ can iteratively revise the response in multiple turns.

After $T$ turns revising, we can obtain the final responses $\{H^T_1, H^T_2, \dots, H^T_N\}$ from each crew agent.
To form the final answer $y$ to user query $x$, the leader agent summarizes all the responses by incorporating the global experiences $E_g$ and its review $R^t_{l}$:
\begin{equation}
    y = A^l(\{H^T_1, H^T_2, \dots, H^T_N\}, E_g, R^t_{l}).
\end{equation}
Next, we employ an LLM-based evaluator agent $P$ to evaluate the quality of the response $y$:
\begin{equation}
    m = P(x, y),\label{eq:eval-rev}
\end{equation}
where $m$ is an overall review which not only contains the judgment of the quality but also illustrates the reason why $P$ gives this evaluation result.
\begin{tcolorbox}[colback=black!1!white,colframe=black!57!white,boxsep=1pt,left=1pt,right=1pt,top=1pt,bottom=1pt]
\instrctionsize
You will be given a multi-day travel plan. The task is \{total\_task\} 

Your task is to evaluate the travel plan on specific criteria.

Please make sure you read and understand these instructions carefully. Please keep this document open while reviewing, and refer to it as needed.

\textbf{Evaluation Criteria:}

1. Plan Customization (1-20): [Evaluation Criteria of P.Cu.]

2. Plan Novelty (1-20): [Evaluation Criteria of P.N.]

3. Plan Correctness (1-20): [Evaluation Criteria of P.Co.]

\textbf{Evaluation Steps:}

1. Review the travel plan in detail.

2. Compare the plan against each criterion.

3. Assign a score for each criterion on a scale of 1 to 20, where 1 is the lowest and 20 is the highest, based on the Evaluation Criteria.

\textbf{Travel Plan:}

\{Travel\_Plan\}
\end{tcolorbox}

\subsection{$360^\circ$ Performance Assessment}\label{sec:360}

In this paper, inspired by the performance assessment of employees in large companies, we propose an agent feedback method, \revm. 
This method comprehensively evaluates the shortcomings of agents in completing tasks through three types of evaluations: \textbf{self-level}, \textbf{peer-level}, and \textbf{supervisor-level}. 
Different from the agent evaluation methods only based on the agent self-reflection~\cite{madaanSelfRefineIterativeRefinement2023}, \revm boosts the mutual collaboration among agents in multi-agent systems.
It enables the evaluation of the agent capabilities from the perspectives of different agents (\aka sub-tasks), thereby encouraging each agent to consider and solve the problem from a global perspective.

After each agent responds to their own sub-task, we will conduct a \revm to assess whether their responses effectively complete the sub-task and whether they are helpful to other agents in solving their tasks. 
Subsequently, this review will be provided to the agent, and the agent will then revise their response to the sub-task based on the review from other agents. 
Through several turns of such iterative evaluations, agents can continuously refine their responses by incorporating feedback from others, thereby improving the quality of their responses.
Since our \revm is a multi-turn generation and assessment paradigm, in this section, we illustrate the details of \revm method by using the agent action in $t$-th turn as an example.

\paratitle{Self-level Assessment.}
Similar to the previous work~\cite{madaanSelfRefineIterativeRefinement2023}, we also employ self-reflection on the response of the agents:
\begin{equation}
    R^t_{s,i} = A^c_i(H^t_i, I^c_i),
\end{equation}
where $R^t_{s,i}$ is the self-reflection of $i$-th agent $A^c_i$ according to its response $H^t_i$ and task instruction $I^c_i$.

\paratitle{Peer-level Assessment.}
In LLM-based multi-agent systems, cooperation among crew agents is prevalent, thus facilitating mutual assistance among crew agents can enhance the overall system performance.
Therefore, we introduce peer-level assessment among crew agents, which allows for assessing the performance of intelligent agents from the perspectives of other tasks, thereby obtaining a more comprehensive evaluation.
Additionally, it can also boost collaboration among the agents.
Specifically, a crew agent reviews the responses of other crew agents and generates feedback for others:
\begin{equation}
    R^t_{i,j} = A^c_i(H^t_i, H^t_j, I^c_j),
\end{equation}
where $R^t_{i,j}$ is the feedback for the response $H^t_j$ generated by $j$-th agent, which is reviewed by the $i$-th agent $A^c_i$.

\paratitle{Supervisory-level Assessment.}
On the other hand, assessment from the leader agent $A^l$ can evaluate the completion of sub-task by agents, and also assist them in better problem-solving from a comprehensive perspective. 
Therefore, we employ the leader agents $A^l$ to evaluate the performance of the crew agent according to its response $H^t_i$ and instruction $I^c_i$:
\begin{equation}
    R^t_{l,i} = A^l(H^t_i, I^c_i),
\end{equation}
where $R^t_{l,i}$ is the review for the $i$-th agent written by the leader agent $A^l$.

After conducting the \revm from three different sources, in the $t$-th turn, the crew agent $A^c_i$ will receive a review set $R^t_{i}$ which contains three types of review:
\begin{equation}
    R^t_{i} = \{R^t_{*,i}, R^t_{l,i}, R^t_{s,i}\}.
\end{equation}
For the leader agent, it only has the self-level assessment review $R^t_{l}$.

\subsection{Dual-level Experience Pool}\label{sec:exp-pool}

For company managers, performance evaluation of employees is not the ultimate goal.
Instead, it is essential to enable employees to accumulate experience from evaluation assessment and perform better in subsequent tasks. 
Inspired by this, we propose the \expm, which involves constructing dual-level experience pools: local and global experience pools, to help each agent and the entire multi-agent system accumulate reusable of problem-solving experience during the task completion process.

\subsubsection{Local-level Experience Construction}

Local experience pertains to each agent itself in completing the current sub-task, summarized from previous multiple turns of \revm in the current task, thereby helping the agent achieve better results in the next turn. 
After obtaining the reviews $R^t_{i}$, the agent $A^c_i$ will reflect their local experience $e^{\prime}_l$:
\begin{equation}
    e^{\prime}_l = A^c_i(H^t_i, R^t_{i}),
\end{equation}
where the crew agent $A^c_i$ is prompted by the following instruction:
\begin{tcolorbox}[colback=black!1!white,colframe=black!57!white,boxsep=1pt,left=1pt,right=1pt,top=1pt,bottom=1pt]
\instrctionsize
Based on feedback from others, past experiences, and from the perspective of one's own role, summarize some experiences that you may use in the future. Your answer should be as concise as possible. Do not output other non output required content.

Your role: \{role\}.

\textbf{Feedback from others:}
\{peer\_feedback\} $R^t_{i}$

\textbf{Previous experience: }
\{pre\_exp\} $E_{l,i}$

Structured output requirements:

Role: [Your Role]

Experience: [Short experience description]
\end{tcolorbox}
Then we append the $e^{\prime}_l$ into the local experience pool $E_{l,i}$ for the $i$-th crew agent $A^c_i$.

\subsubsection{Global-level Experience Construction}

Global experience refers to reusable experiences for the entire multi-agent system, and we build these experiences from the final response of each crew agents and the feedback $m$ of the evaluator $P$ after completing the task:
\begin{equation}
    e^{\prime}_g = A^l(m, \{H^T_1, H^T_2, \dots, H^T_N\}),\label{eq:global-exp}
\end{equation}
where the leader agent $A^l$ is prompted by the following instruction:
\begin{tcolorbox}[colback=black!1!white,colframe=black!57!white,boxsep=1pt,left=1pt,right=1pt,top=1pt,bottom=1pt]
\instrctionsize
Obtain new experiences based on previous experiences and current scores. The new experience should be summarized while presenting the experience gained from this task.
Here are the final results: \{Final\_Res\} ${H}$

Here are the evaluation of the task: \{evaluation\} ${m}$

As a leader, you need to learn experience from this task, and the format should be "\textbf{Where did I do well this time}... \textbf{why didn't I do well this time}... \textbf{next time I should}...".Note that these experiences are for this type of task. The answer should be as concise as possible.
\end{tcolorbox}
Finally, we can accumulate the global experience pool by appending the $e^{\prime}_g$ into the local experience pool $E_g$.

%% file: sections/05-experiment.tex
\section{Experimental Setup}~\label{sec:experiment}

\subsection{Tasks and Datasets}

Following existing works~\cite{wangUnleashingCognitiveSynergy2023,xieTravelPlannerBenchmarkRealWorld2024}, we verify our \model performance on two tasks, creative writing, and travel plan making.
\begin{enumerate*}[label=(\roman*)]
\item \textbf{Creative Writing}. We use \textbf{\texttt{Trivia Creative Writing}}~\cite{wangUnleashingCognitiveSynergy2023} dataset, which contains 100 samples. Each sample is required to generate a story to cover the required input topic and input questions.
\item \textbf{Travel Plan Making}. We use \textbf{\texttt{Trip Planning}}~\cite{xieTravelPlannerBenchmarkRealWorld2024} dataset that comprises travel plans for 50 world-famous attractions, with specific requirements generated based on the characteristics of the destinations. It includes the following fields: destination, basic requirements, duration of travel, season, type of travelers, number of travelers, and budget.
\end{enumerate*}

\subsection{Evaluation Metrics}
Since the objectives of travel planning tasks and creative writing tasks are not the same, we employ different evaluation perspectives in our experiments to assess the quality of the model-generated results. 

For the story generation task, we employ the metric matched answer rate $Match\%$ \textit{(M\%)} used in \citet{wangUnleashingCognitiveSynergy2023}, to measure the percentage of generated story cover the ground truth target answers for each question. 
We also propose two semantic evaluation metrics, which prompt the LLM to evaluate whether the generated results satisfy the specific aspects including: (1) \textit{Emotional Engagement (E.E.)}, whether the generated story evokes the reader's emotion and empathy; (2) \textit{Insightfulness (Ins)}, whether the story carries the insightful plot and cause a profound impact on readers.

For the travel planning task, we also employ three semantic metrics for evaluating three aspects: (1) \textit{Plan Correctness (P.Co.)}, whether the plan covers all required details and the plan is reasonable; (2) \textit{Plan Novelty (P.N.)}, whether the plan is novel and creative; (3) \textit{Plan Customization (P.Cu.)}, whether the plan is customized to the personal interests and preferences of traveler.

For the logic grid puzzle task, we use one evaluation metric:
\textit{Accuracy (Acc)}, which measures the percentage of correctly predicted house numbers compared to the ground truth targets.














\subsection{Baselines} 




We compare our method with agent-based methods, including:
\begin{enumerate*}[label=(\roman*)]
    \item \texttt{GPT-4}~\citep{openaiGPT4TechnicalReport2023} is a closed-source LLM from OpenAI. We conduct task solving as an in-context learning that uses a single instruction with the same input data as our \model.
    \item \texttt{Mistral}~\citep{jiangMistral7B2023} is an open-source LLM and we use the mixture-of-experts version named Mistral-medium with 46 billion parameters, and use the same prompt as \texttt{GPT-4}.
    \item \texttt{OKR}~\citep{zhengAgentsMeetOKR2023} is a hierarchical multi-agents collaboration framework. They propose the hierarchical OKR generation to decompose objects into multiple sub-objects and assign new agents based on key results and agent responsibilities.
    \item \texttt{AgentVerse}~\citep{chenAgentVerseFacilitatingMultiAgent2023} is a multi-agent framework that enhances collaborative problem-solving by employing specialized agents working together through structured interactions.
    \item \texttt{SPP}~\citep{wangUnleashingCognitiveSynergy2023} propose the Solo Performance Prompting (SPP), which transforms a single LLM into a cognitive synergist by engaging in multi-turn self-collaboration with multiple personas.
\end{enumerate*}

We implement the \texttt{OKR},\texttt{AgentVerse} and \texttt{SPP} with the backbone LLM \texttt{GPT-4}.
We also employ two variants of \model: \textbf{\model-G} and \textbf{\model-M} with \texttt{GPT-4} and \texttt{Mistral} as the backbone respectively. 

\subsection{Implementation Details}

We conduct 2 turns of $360^\circ$ assessment.
We use 1 leader agent and set the number of crews is range from 3 to 5, we ask the LLM to follow that range by stating the range in prompt text. 
The number of experiences selected by the leader from the global-level experience pool is 10. 
We use GPT-4 version of 1106-preview with a temperature of 1.0. 
Our anonymized source code is publicly released \footnote{https://anonymous.4open.science/r/360-REA-0FBD}.
















%% file: sections/06-result-and-analysis.tex
\section{Experimental Results}~\label{sec:result}

\begin{table*}[htbp]
\centering
\small

\begin{tabular}{lccc|ccc}
\toprule
\multirow{2}{*}{\textbf{Model}} & \multicolumn{3}{c|}{Creative Writing} & \multicolumn{3}{c}{Travel Plan Making} \\
 & \textbf{M\%}($\uparrow$) & \textbf{E.E.}($\uparrow$) & \textbf{Ins}($\uparrow$) & \textbf{P.Co.}($\uparrow$) & \textbf{P.N.}($\uparrow$) & \textbf{P.Cu.}($\uparrow$) \\
\midrule
\texttt{GPT-4}~\cite{brownLanguageModelsAre}               & 77.1 & 63.2 & 66.3 & 83.1 & 70.2 & 93.3 \\
\texttt{Mistral}~\cite{jiangMistral7B2023}      & 62.1 & 55.4 & 60.5 & 82.3 & 66.3 & 92.4 \\
\texttt{OKR}~\cite{zhengAgentsMeetOKR2023}                 & 54.5 & 58.3   & 62.8 & 86.4 & 75.7 & 93.0   \\
\texttt{AgentVerse}~\cite{chenAgentVerseFacilitatingMultiAgent2023}                 & 75.8 & 58.6 & 54.7   & 85.6 & 75.9 & 94.1 \\
\texttt{SPP}~\cite{wangUnleashingCognitiveSynergy2023}                 & 83.8 & 54.2 & 57.0   & 88.8 & 74.4 & 92.6 \\
\midrule
\texttt{\model-G}  & \textbf{87.0} & \textbf{66.1} & \textbf{72.8} & \textbf{89.6} & \textbf{80.2} & \textbf{94.2} \\
\texttt{\model-M} & 71.2 & 59.6 & 65.1 & 84.9 & 70.4 & 94.0 \\
\texttt{\model-G w/o ExpPool}  & 83.1 & 63.1 & 66.2 & 88.0   & 76.8 & 93.7 \\
\texttt{\model-G w/o $360^\circ$F} & 84.1 & 63.9 & 67.1 & 88.6 & 76.6   & 93.9 \\
\bottomrule
\end{tabular}
\caption{The overall performance of \model comparing with baselines and ablation studies on the two tasks.}
\label{tab:main_result}
\end{table*}

\subsection{Overall Performance}
Table~\ref{tab:main_result} shows the performance of our proposed \model and baselines in terms of three metrics on two benchmark datasets.
We can find that \model shows superior performance in terms of all metrics compared to the state-of-the-art method \texttt{SPP} which is also a multi-agent hierarchical collaboration framework with agent discussion.
This phenomenon indicates that our proposed $360^\circ$ assessment and dual-level experience pool can help the agents to generate better responses in a higher quality. 
And we can also find that each variant of our \model (\eg \model-G and \model-M) achieves better performance than simply prompting its LLM backbone, it demonstrates our proposed framework can adapt to any LLM backbones.



\subsection{Human Evaluation}
As using LLM as the evaluator is sometimes not robust, we also employ human evaluation to further validate the effectiveness of the model-generated answers.
For creative writing tasks, we evaluate the quality of the generated response in terms of: (1) \textbf{Coherence}, the score indicating the story's coherence; (2) \textbf{Creative}, the score indicating the story's creativeness. 
For travel plan making, we evaluate the response in two aspects: (1) \textbf{Correct}, the score showing whether the plan is correct and complete, and (2) \textbf{Customized}, the score measuring whether the plan is customized. 
We hire 3 educated annotators with Bachelor degree to conduct double-blind annotation on randomly shuffled results. 
The annotating scores range from 1 to 100.
From Table~\ref{tab:human}, we observed that our \model excels in all aspects compared to the strong multi-agent baseline \texttt{SPP} and \texttt{GPT-4}, which are the strongest baseline in the two tasks (Table~\ref{tab:main_result}). We also conclude that the automatic metrics in Table~\ref{tab:main_result} are also consistent with the human evaluation.



\begin{table}[htbp]
\centering

\resizebox{1.0\columnwidth}{!}{
\begin{tabular}{lcc|cc}
\toprule
\multirow{2}{*}{\textbf{Model}} & \multicolumn{2}{c|}{Creative Writing} & \multicolumn{2}{c}{Travel Plan Making} \\
\cmidrule{2-3} \cmidrule{4-5}
 & Coherence($\uparrow$) & Creative($\uparrow$) & Correct ($\uparrow$) & Customized ($\uparrow$) \\
\midrule
\texttt{GPT-4} & 76.2 & 76.1 & 72.6 & 74.3 \\
\texttt{SPP} & 73.8 & 74.0 & 74.5 & 77.5 \\
\texttt{\model-G} & \textbf{79.1} & \textbf{79.2} & \textbf{79.8} & \textbf{84.4} \\
\bottomrule
\end{tabular}
}
\caption{Results of human evaluation on two tasks.}
\label{tab:human}
\end{table}

\subsection{Ablation Study}
To verify the effectiveness of each module of our \model, we conduct ablation studies with model \model-G with several ablation variants:
\begin{enumerate*}[label=(\roman*)]
\item \textbf{\texttt{\model-G w/o ExpPool}}: We remove both local and global experience from \model and directly use the assessment from other agents as the prompt of each agent.
\item \textbf{\texttt{\model-G w/o $360^\circ$F}}: We directly use the output of each crew agent without conducting all levels of assessment.
\end{enumerate*}


As the results shown in Table~\ref{tab:main_result}, we found that the \texttt{\model-G w/o ExpPool} method achieved lower scores compared to our full model at most metrics, indicating the necessity of using experience pool to recall reusable experience. 
And \texttt{\model-G w/o $360^\circ$F} mainly underperforms \model-G, showing the effectiveness of conducting all level assessments among multi-agents.

\begin{table}[htbp]
\centering

\resizebox{1.0\columnwidth}{!}{
\begin{tabular}{lccc}
\toprule
\textbf{Model} & \textbf{M\%} & \textbf{E.E.} & \textbf{Ins} \\
\midrule
\texttt{\model-G}  & \textbf{87.0} & \textbf{66.1} & \textbf{72.8} \\
\texttt{\model-G w/o GlobalExp} & 84.0 & 64.0 & 68.5 \\
\texttt{\model-G w/o LocalExp} & 83.7 & 65.2 & 69.9 \\
\bottomrule
\end{tabular}
}
\caption{Comparison on different usages of experience pools.}
\label{tab:exp-pool}
\end{table}

\begin{table}[htbp]
\centering

\resizebox{1.0\columnwidth}{!}{
\begin{tabular}{lccc}
\toprule
\textbf{Model} & \textbf{M\%} & \textbf{E.E.} & \textbf{Ins} \\
\midrule
\texttt{\model-G}  & \textbf{87.0} & \textbf{66.1} & \textbf{72.8} \\
\texttt{\model-G w/o Supervisory}  & 85.1 & 65.9 & 71.9 \\
\texttt{\model-G w/o Peer} & 84.8 & 64.0 & 67.5 \\
\texttt{\model-G w/o Self} & 86.2 & 64.6 & 68.7 \\
\bottomrule
\end{tabular}
}
\caption{Performance of ablation models on different assessment levels.}
\label{tab:diff-360}
\end{table}

\subsection{Case Study}

Table~\ref{fig:case} shows an example of creative writing generated by \texttt{SPP} and \model-G.
The \textbf{bold text} in the response indicates the correct answer to the input question. 
\model-G's results contain the answers to all five questions, whereas \texttt{SPP} only correctly answered two of them.
The text in \textcolor{green}{green} and \textcolor{red}{red} indicates the strengths and weaknesses in the generated story respectively. 
The green parts show \model-G's output has a rich plot and illustrates the cause and ending of the story. 
However, the red parts show \texttt{SPP}'s story plot lacks a brief introduction and the use of ``Valerie'' is abrupt and weird. 
From this case, we observed that \model-G obtains more reasonable, logical, and fluent stories rather than \texttt{SPP}.
We also show an example with the corresponding experiences in the Appendix~\ref{sec:appendix}


\begin{table} 

  \centering
  \scriptsize
  \resizebox{1.0\columnwidth}{!}{
  \begin{tabular}{p{0.59\columnwidth}|p{0.41\columnwidth}}
    \toprule
    \multicolumn{2}{c}{\textbf{Task Definition}} \\
    \midrule
    \multicolumn{2}{p{1.05\columnwidth}}{Write a short and coherent story about Legend of Zelda that incorporates the answers to the following 5 questions: Which British monarch famously said ‘I don’t wish to open windows into men’s souls’? Which British singer played Jareth the Goblin King in the 1986 film ‘Labyrinth’? The 1987 film ‘Cry Freedom’ is a biographical drama about which South African civil rights leader? Which British actress played Valerie in the 1997 film ‘Nil By Mouth’? What was the name of the Brazilian player who won the Women’s Singles Finals at Wimbledon in 1959, 1960 and 1964?} \\
    \midrule
    \textbf{\model Output} & \textbf{SPP Output} \\
    \midrule
In the age of legends and mystical chants, the kingdom of Hyrule \textcolor{green}{stood on the precipice of an age-old challenge, the Enchanted Labyrinth.} Queen Zelda, a wise and fair ruler whose principles mirrored those of Queen \textbf{Elizabeth I}, ...

Under a sky lit with stardust, Link, the hero garbed in green, stood before the labyrinth's entrance, his heart bracing for the trials ahead. A figure emerged from the shadows, a labyrinth guardian as enigmatic and charismatic as Jareth the Goblin King, whose likeness to the British singer \textbf{David Bowie} was unmistakable, ...

\textcolor{green}{Deeper within the winding passages}, Link crossed paths with a spirit named \textbf{Steve Biko}, ...

Throughout the labyrinth, the presence of Valerie, inspired by \textbf{Kathy Burk}e's portrayal ...

\textcolor{green}{In the labyrinth's final sanctum}, a guardian awaited, her grace and skill rivaling that of \textbf{Maria Bueno}, ...

\textcolor{green}{The kingdom rejoiced as Link returned from the labyrinth}...

Thus, the saga of the Enchanted Labyrinth of Hyrule would forever resonate through the ages—a testament to the power of knowledge, the resonance of culture, and the eternal dance of history intertwining with the threads of the present.

& In the mystical land of Hyrule, a legend unfolded beneath the watchful eyes of the Triforce. Once there was a queen, wise and just, known as \textbf{Elizabeth}, ...

Out from this labyrinth emerged an enigmatic figure, Jareth, with a voice as captivating as the British singer of old, \textbf{David Bowie} ...

But destiny had chosen a hero, young Link, to rise against this new foe. ...

\textcolor{red}{In his company was the valiant Valerie}, who resembled the strong-willed British actress seen in 'Nil By Mouth'. Together they braved the challenges of Jareth's labyrinth, finding strength in unity.

Throughout their quest, they were inspired by the tales of Maria, ...

\textcolor{red}{After a long and arduous journey}, Link and his companions faced Jareth and freed Hyrule from the labyrinth's spell. They returned to their queen, ...\\
    \bottomrule
  \end{tabular}
  }
\caption{Examples of the generated response of two methods on the creative writing task.}
\label{fig:case}
\end{table}

\section{Discussion}~\label{sec:discussion}

\subsection{Analysis on Different Level Experiences}

In this paper, we propose a novel dual-level experience pool. 
In this section, we employ two ablation models to explore which layer of the experience pool is more effective and conduct experiments on the creative writing dataset. 
The results in Table~\ref{tab:exp-pool} indicate that local experience is the most helpful. 
One possible reason for this phenomenon is that local experience contains lower-level experiences, including more specific task-completion skills. 
From Table~\ref{tab:exp-pool}, we also observe that compared to the \model, the model without global experience has a performance decrease of 5.9\% in terms of Ins, indicating that the higher-level experience for task-solving in global experience also plays a crucial role in improving model performance.


\subsection{Analysis on Different Level Assessment}

We also investigate the effectiveness of different levels of assessment on the creative writing task.
From Table~\ref{tab:diff-360}, we find that all the ablation models achieve worse performance than \model-G which demonstrates the effectiveness of each level of assessment in \model.
We also find that the variant model without the peer-level assessment achieves the worst among all the ablation models.
This phenomenon indicates that peer-level assessment provides the greatest improvement in overall performance, confirming our hypothesis that mutual evaluation among crew members helps each agent not only in completing their task but also in assisting other agents as much as possible.


%% file: sections/08-conclusion.tex
\section{Conclusion}\label{sec:conclusion}

In this paper, we propose a hierarchical multi-agent framework, \fullmodel (\model), which represents a significant step towards enhancing the capabilities of LLM-based agents in tackling complex tasks. 
Unlike existing approaches that primarily focus on team optimization or self-reflection, \model incorporates a comprehensive agent performance evaluation strategy through the novel \revm method. 
This method provides multi-perspective and fine-grained feedback to agents, in which agents receive valuable insights from self, peers, and supervisory levels. 
Furthermore, to substantively improve agent task-solving capabilities, we introduce the \expm for the accumulation of experiences based on evaluation feedback. 
Our experiments on complex task datasets demonstrate the effectiveness of \model in achieving superior performance compared to the state-of-the-art baselines. 
By emphasizing the importance of comprehensive evaluation and experience accumulation, \model establishes a promising direction for advancing the field of multi-agent systems built on LLMs, offering a practical and impactful framework for addressing intricate tasks.

\clearpage

\section*{Limitations}\label{sec:lmitation}

In this paper, we only use the text as the input.
However, multi-modal data (\eg images and videos) is commonly used in many tasks.
Since existing multi-modal LLMs are capable of understanding both textual and multi-modal information, our multi-agent framework can adapt to leverage multi-modal inputs in many complex tasks. 
In the future, we plan to incorporate multi-modal information into our framework.

\section*{Ethics Statement}

The multi-agent collaboration framework proposed in this paper is built upon LLM which is trained by reinforcement learning with human feedback (RLHF) \cite{lambert2022illustrating}. 
Since LLM has already been aligned with human values and preferences during the RLHF stage, the likelihood of generating unethical content is low. 
However, existing LLMs still cannot completely prevent generating harmful content. 
Therefore, the answers generated by the multi-agent collaboration framework still require further human expert verification before use.

\section*{Acknowledgements}\label{sec:acknowledgement}

This work was supported by the Natural Science Foundation of Shandong Province (ZR2023QF159), the National Natural Science Foundation of China (U2001212, 62032001, 61932004, 62306330, and WDZC20235250103), and Young Elite Scientist Sponsorship Program by CAST (2023QNRC001) under Grant No. YESS20230367.

%% file: sections/09-appendix.tex
\section{Example of Generated Response}~\label{sec:appendix}

In this section, we show an example of the generated response when training our \model in the early training stage (with few experiences in the global experience pool).
We show the output of several modules in \model including the review $m$ of the evaluator (shown in Equation~\ref{eq:eval-rev}), and the global experience $e_g^{\prime}$ (shown in Equation~\ref{eq:global-exp}) learned from this example.

In the review $m$ of the evaluator, we can see that our generated travel plan still has some flaws, with a lower score in terms of feasibility, and the evaluator has provided corresponding reasons. 
The text in \textcolor{blue}{blue} indicates the flaw mentioned in the evaluation.

In the global experience $e_g^{\prime}$, the text in \textcolor{blue}{blue} indicates that \model summarizes some detailed skills from the review $m$.
The text in \textcolor{green}{green} indicates the experiences that are summarized from input experiences when generating this response (\aka experiences obtained from previous samples).
It demonstrates that our \model can not only summarize new experiences from the current task-solving process, but also keep previous experiences in mind and form new experiences from them,

\textbf{Task description:}

destination:Barcelona, Spain

days:6

description:A 6-day adventure in Barcelona, experiencing its unique architecture by Gaudí, vibrant food markets, and beautiful beaches.

season:Summer

month:June

interests:Architecture, Food Markets, Beaches

members:adults:4, children:0

preferences:Cultural Explorers

Budget Range:Mid-range

\textbf{The generated travel plan by \model:}

Day 1: Arrival and Architectural Introduction

- Arrive at Barcelona and check-in at Hotel Midmost.

- Afternoon: Private walking tour with a local architect to explore lesser-known Modernista sites.

- Evening: Guided visit to Recinte Modernista de Sant Pau.\\

Day 2: Iconic Gaudí and Culinary Delights

- Morning: Guided tour of La Sagrada Familia.

- Afternoon: La Boqueria market tour followed by a cooking class with a local gastronomy expert.

- Evening: Free time for relaxation or independent exploration.\\

Day 3: Park Güell Wonders and Coastal Activities

- Morning: Park Güell visit including a creative workshop on Gaudí's nature inspirations.

- Afternoon: Barceloneta Beach with sunbed reservation and option for a Kite-surfing lesson or beach volleyball.

- Evening: Attend a local jazz festival by the sea or a Festa Major street festival.\\

Day 4: Traditional Markets and Twilight Art

- Morning: Local market tour at Mercat de la Libertat.

- Afternoon: Casa Vicens visit and leisurely stroll through La Ribera district to see Basilica de Santa Maria del Mar.

- Evening: Visit to lesser-known architectural wonders with a rooftop concert featuring Spanish guitar at Torre Bellesguard.\\

Day 5: Artistic Immersion and Seaside Farewell

- Morning: Beach clean-up activity followed by a picnic or if weather permits, a visit to Torre Bellesguard including a rooftop concert.

- Afternoon: Artistic workshop inspired by Gaudí.

- Evening: Passeig de Gràcia tour with Casa Batlló \& La Pedrera and reserved seating at a seaside restaurant for a farewell dinner.\\

Day 6: Leisure and Departure

- Morning: Guided beach yoga session or another family-friendly activity such as a visit to Hospital de Sant Pau’s pavilions.

- Afternoon: Free for last-minute explorations or shopping.

- Airport transfer via pre-booked minivan service.

Emergency Strategy and Contingencies:

- Daily travel insurance briefings, weather advisories, and emergency point of contact sharing.

- Accessibility to VIP access to indoor venues like the Picasso Museum in case of hot weather or heavy rain.

- Communication protocols set to inform the group of any changes.\\

Logistics and Budget Management:

- Hola BCN Travel Card and bicycle rentals for daily commutes.

- Real-time budget tracking with contingency fund for spontaneous decisions.

- Comprehensive local interactions from curated market tours to hands-on workshops, ensuring rich cultural exposure.\\

Concluding Remarks:

This plan strikes a balance between gourmet experiences, art and architecture, and coastal leisure, aptly tailored for adults and takes into account Barcelona's summer climate. Each activity is selected to enhance the group's travel experience, seasoned with local insights and flexibility for a safe and memorable journey in Barcelona.\\

\textbf{Review from evaluator:}

Alignment with Interests and Preferences: 20

The travel plan meticulously aligns with the group's interests in architecture, food markets, and beaches, as well as their preference for cultural exploration. Activities such as private architectural tours, cooking classes, and creative workshops directly cater to these interests. The inclusion of iconic Gaudí sites, local markets, and beach activities ensures a tailored experience for the group's cultural and leisure preferences.

Variety and Balance: 19

This itinerary offers a commendable balance of activities, from architectural tours and artistic workshops to leisure time at the beach and culinary experiences. The diversity of experiences ensures engagement throughout the trip, catering to both cultural interests and relaxation needs. The inclusion of evening activities like rooftop concerts and a farewell dinner at a seaside restaurant adds a delightful variety to the travel experience.

Feasibility and Logistics: 17

Overall, the plan appears feasible with a coherent schedule and consideration for logistics, including the use of the Hola BCN Travel Card and bicycle rentals for daily commutes. \textcolor{blue}{However, the packed itinerary might challenge the group's energy levels, especially with multiple activities scheduled on the same day.} Ensuring sufficient rest and travel time between activities would enhance the feasibility of this ambitious schedule.

Local Insights and Unique Experiences: 20

The plan excels in offering unique, local experiences, from a market tour followed by a cooking class with a local expert to creative workshops inspired by Gaudí. These activities not only provide an immersive cultural experience but also allow for meaningful interaction with the local community. The thoughtful inclusion of lesser-known sites and local festivals further enriches the travel experience with authentic insights into Barcelona's culture and lifestyle.

Overall Evaluation:

This travel plan is excellently crafted, demonstrating a deep understanding of the group's preferences and a strong alignment with their cultural and leisure interests. While there is a minor consideration for the feasibility of the packed schedule, the variety of activities and the emphasis on local, unique experiences promise a memorable and enriching visit to Barcelona. The plan's thoughtful contingencies and logistics management also underscore its practicality and attentiveness to the group's comfort and experience.\\

\textbf{Generated Global Experience:}\\

Where did I do well this time: I excelled in aligning the travel plan closely with the group's specified interests and preferences, especially in architecture, food markets, and beaches. My success in integrating unique and authentic local experiences, such as creative workshops and cooking classes with local experts, significantly enriched the cultural exploration aspect of the trip.

Why didn't I do well this time: \textcolor{blue}{I underestimated the importance of ensuring adequate rest periods and manageable schedules within the plan.} The feasibility and logistics score reflected this oversight, indicating that the itinerary might be too packed, potentially leading to fatigue and limiting the group's ability to fully enjoy each activity.

Next time I should: Focus on creating a more balanced itinerary that not only aligns with the travelers' interests \textcolor{blue}{but also accounts for practical aspects such as rest, travel times between activities, and the overall pace of the trip.} This means potentially reducing the number of activities per day or incorporating more flexible, free time slots to ensure a comfortable and enjoyable experience for all travelers. \textcolor{green}{Additionally, I should consider more deeply the seasonal appropriateness of each activity, ensuring that the group is comfortable and able to enjoy the trip regardless of weather conditions.}